# Deep Latent Emotion Network for Multi-Task Learning

Huangbin Zhang, Chong Zhao, Yu Zhang, Danlei Wang, Haichao Yang

Tencent PCG, China

{tobynzhang, chozhao, edmondzhang, danleiwang, haichaoyang}@tencent.com

## ABSTRACT

Feed recommendation models are widely adopted by numerous feed platforms to encourage users to explore the contents they are interested in. However, most of the current research simply focus on targeting user's preference and lack in-depth study of avoiding objectionable contents to be frequently recommended, which is a common reason that let user detest. To address this issue, we propose a Deep Latent Emotion Network (DLEN) model to extract latent probability of a user preferring a feed by modeling multiple targets with semi-supervised learning. With this method, the conflicts of different targets are successfully reduced in the training phase, which improves the training accuracy of each target effectively. Besides, by adding this latent state of user emotion to multi-target fusion, the model is capable of decreasing the probability to recommend objectionable contents to improve user retention and stay time during online testing phase. DLEN is deployed on a real-world multi-task feed recommendation scenario of Tencent QQ-Small-World with a dataset containing over a billion samples, and it exhibits a significant performance advantage over the SOTA MTL model in offline evaluation, together with a considerable increase by 3.02% in view-count and 2.63% in user stay-time in production. Complementary offline experiments of DLEN model on a public dataset also repeat improvements in various scenarios. At present, DLEN model has been successfully deployed in Tencent's feed recommendation system.

## CCS CONCEPTS

• **Information systems** → **Recommender systems; Retrieval models and ranking** • **Computing methodologies** → **Multi-task learning;** Neural networks;

## KEYWORDS

Multi-task Learning, Recommender System, Bayesian Probability, Latent Emotion

## 1 INTRODUCTION

In recent years, with the development of mobile applications, the information flow business has gained huge growth, in which the personalized recommendation system plays an important role. Unlike traditional ad recommendation algorithms, feed recommendation usually has to optimize several different targets at the same time. The purpose of personalized recommendation systems is to recommend content to users that they might prefer based on their interaction behavior, and to attract them to consume more content by minimizing their objectionable level, thus realizing business value.

However, being different from ad recommendations, which mostly focus on optimizing the click-through rate, the goals to be optimized in a feed recommendation system usually include click-through rate, like rate, push rate, favor rate, comment rate, and follow rate, etc. The overall metrics relevant to the business value of the infomercial business ultimately are the number of Page Views (PV) and use stay time.

Lately, MTL models have been found to be well suited for feed recommendation [4-8]. Compared with single-objective optimization models, neural network based MTL models can share low-level information among different tasks and obtain overall benefit by linking them. However, the existing MTL model is not sufficiently developed to cope with the current feed scenario.

In most feed recommendation scenarios, user's interaction behaviors are relatively sparse, and the number of negative samples for each target is usually much larger than the number of positive samples from a point-wise perspective. When different positive feedback behaviors (such as like, follow, push, comment) are available, we can observe the following two phenomena:

>    1) Users usually select only one or a few positive feedback behaviors to show their preference to a feed, which leads to a result in one feed interested by users being a positive sample on some targets while negative on most others.
>    2) The interaction behavior habits of different users are greatly varied. When expressing preference to the same feed, some users prefer to click like button, while some tend to share it, etc.

These two phenomena lead to the conflict of sharing parameters in the shallow layer of the network when the MTL model is learning.

Based on the above phenomena, we propose a Deep Latent Emotion Network (DLEN) model, which constructs the user's latent emotional tendency towards a feed as a hidden target. The other individual explicit behavior targets and this hidden target are coupled by Bayesian probability formula and trained together. Compared with existing multi-task models such as MMOE[6] and PLE[8], which decrease the training conflict among different targets by sharing or segregating between upper layer networks, DLEN reduces the conflicts by sharing of targets to improve the training effectiveness. Besides, the hidden target can be used as part of online multi-target fusion to model the probability of user preferring a feed, so as to increase the Page Views (PV) and Stay Time by avoiding recommending feeds that user do not prefer.

To evaluate the effectiveness of the DLEN model, we trained this model on Tencent's real business dataset and open dataset and compared the results with base model MMOE. The DLEN model result manifests that both offline and online metrics are significantly improved.

The main contributions of this paper include the following points:

>    1. Under a multi-task recommendation scenario, we use Bayesian tools to express the user's latent state (the probability of the user prefer a feed) explicitly through mathematical formulas and to promote model convergence with rigorous assumptions.
>    2. Based on the proposed latent state, a new multi-task model architecture and its training scheme are designed, which utilizes user's explicit interaction samples to infer the probability of users preferring a feed with semi-supervised learning.
>    3. The experiment comparison is conducted on Tencent online dataset and open-source offline dataset. In the offline evaluation, DLEN model can significantly improve the learning effectiveness. Online result demonstrates that involving latent state as model's task in the weighting can explicitly control the metrics of user's stay time and retention rate.

## 2 RELATED WORK

### 2.1 MTL Models for Recommendation

Real-world recommendation scenarios are quite rich in user behaviors, and MTL model can make better use of different user behaviors for modeling their preference. Hence, MTL model has gradually become the mainstream ranking model in recent years.

Some prior researches use matrix factorization based on multi-task model to improve the interpretability of recommendations. Lu et al.[1] proposes a joint learning method that performs rating prediction and recommendation explanation by combining matrix factorization. Similarly, Wang et al.[2] applies a joint tensor factorization to integrate companion learning tasks of user preference modeling for recommendation and opinionated content modeling for explanation. However, limited by the simple model structure and single source of features, the matrix factorization-based models are not sufficient to fully characterize user's preferences.

Taking advantages of neural networks, Caruana et al.[3] proposes a widely used MTL method with a hard parameter sharing structure, in which a shared latent layer is trained in parallel on all the tasks. Trapit et al.[19] also adopts multi-task GRU for scientific paper recommendation. The hard parameter sharing structure is supposed to help each task learn better using what learned by other tasks, but the natural differences between each task may lead conflicts in learning shared representation. Furthermore, ESMM[4] puts forward a new perspective of multi-task modeling with sequential dependence to cope with the challenges of extreme data sparsity and sample selection bias, which however is not suitable for parallel multi-task learning. PAL[18] utilizes a two-tasks model to eliminate position-bias in ads recommendation, which separates the probability of seen and click-through rate.

Up to now, the state-of-the-art MTL models take use of mixture-of-experts and its improved architecture to alleviate learning conflicts caused by differences among tasks. MOE[5] isolates the learning of shared representation in the model somewhat through multiple shared expert components as well as a gate component. On top of this, MMOE[6] makes the gate component exclusive to each task, providing flexibility for the selection of shallow features of each task. Moreover, MRAN[7] utilizes attention-based relationship learning modules to exploit the multi-type relationships between feature and task to guide model learning. Rather than the gates-exclusive method of MMOE, PLE[8] proposes a novel experts-exclusive method and explicitly separates task-common and task-specific expert parameters in the model to avoid conflicts caused by differences between complex learning tasks. However, the existing researches still lack the thorough exploration of the conflict nature of shared parameters, and there is no in-depth analysis of the incompatibility and synergy of each task. Prior works [3,20,21] try to measure the differences between tasks according to the assumption of the data generation process, but in real world the data generation process is often more complicated and hard to measure. The DLEN model we propose in this paper aims to fill in these gaps by modeling the user's latent emotion state with Bayesian probability, and decomposes the label of each task to essentially eliminate the parameter learning conflicts generated by the difference among tasks.

## 2.2 Negative sample analysis

Content exposures that are not converted into interactions are usually defined as negative samples in feed recommendion models. Since the interaction habits of different users vary greatly, the huge number of negative samples on each task contains different emotions of users, such as dislike, apathetic, acceptable, or even like. This procedure brings a bias to the conventional negative sample selection, that is, not all negative samples represent users' disgusting emotions. How to process these negative samples more scientifically in training has been a neglected topic for a long term.

Since in most recommendation scenarios, the user's positive interaction behavior is very sparse, the selection of positive samples is relatively straightforward, while negative samples are often down-sampled at a certain percentage to ensure that there is no huge tilt between positive and negative samples during training[17]. At present, the work related to negative samples in the recommendation model is mainly used to speed up the training of the model through some sampling methods [9,10].

Some prior study was also conducted to discover and process the bias in feedback data, which however focuses on the MNAR(missing not at random) phenomenon on the rating data[11]. Marlin et al.[12] conducted a user survey to show the existence of selection bias in the rating data. [13,14] show the distribution of observed rating data is different from the distribution of all ratings. [12,13,15,16] treat the problem as missing data imputation and model the generative process of rating values and the missing mechanism. Existing work on the research of bias in negative samples is almost blank, so we proposed to introduce Bayesian probability to neural-based model to solve the problem.

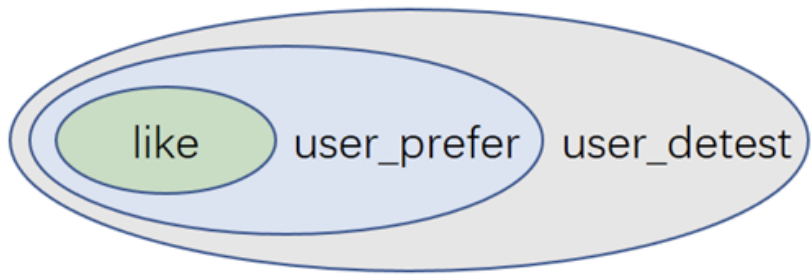

Figure 1: User emotion tendency in samples

In the feed recommendation system, we have noticed two phenomena:
- The probability of user interaction is relatively low, and negative samples account for the vast majority.
- Users' disgusting emotions tend to appear when objectionable contents are frequently recommended.

We further divide the positive and negative samples. Negative samples can be divided into feeds that users detest and feeds that users don't detest but have no interaction. Therefore, in the learning process of the MTL model, it is difficult to make full use of negative samples without distinguishing different types of negative samples. Take the goal of liking as an example, from Figure 1 we can see the relationship between samples.

How to make further efficient use of negative samples and reduce the possibility of recommending unpleasant feeds to users is the key issue for improving user retention. To this end, we propose a user potential state extraction model to make full use of negative samples.

Based on the analysis of user emotional tendency, we further investigated the negative samples of users who like this task. We assume that negative samples can be divided into these three types:

1. The user does not have a "like" behavior, but other positive behaviors occurred.

2. The user does not have any positive behavior, but the user has a positive emotional tendency towards this feed.

3. The user detests this feed.

The existing modeling methods confuse these three situations. Obviously, the information hidden by these three types of negative samples is different. How to further dig out the information hidden by negative samples and reduce the overall disgust of users when experiencing business is a key issue to improve business indicators.

Inspired by this in-depth thinking, we propose a Deep Latent Emotion Network(DLEN) model, which can improve the overall recommendation effect by fully mining the hidden information of negative samples.

## 3 DEEP LATENT EMOTION NETWORK (DLEN) MODEL

Based on observations of user browsing behaviors, we discover that when different users are attracted to the same feed, the type of their interaction might be various. For example, if user A prefer a feed and then click the "like" button, then this sample is a negative sample for the task of "following". Similarly, another user B express interest to the same feed with an action of following the poster, which would be a negative sample for the "like" task. Both users share a same motivation that a user prefer the feed, but they take different behaviors simply because of different interaction habits.

The DLEN model we proposed uses the Bayesian full formula to model the user's emotional tendency toward the feed as a hidden target of MTL. Here are two advantages:

● By increasing the sharing of information between different targets, the learning efficiency of each target is improved while reducing conflicts between targets;

●The latent target is fused with multiple tasks and finally applied to the ranking score. By reducing recommendation of feed that users' disgusting emotions, the user stay time, Page Views (PV), and interaction rate have increased.

### 3.1 Bayesian probability decomposition

We introduced Bayesian probability analysis to characterize the consumption behavior of users on the feed flow. The following is an example of the task of "click-button-to-like".

Using Bayesian formula to expand the probability of positive samples in the task of "click-button-to-like", we have formula:

$$P(CBL) = P(CBL|UP) * P(UP) + P(CBL|\overline{UP}) * P(\overline{UP}) \quad (1)$$
$$P(UP) + P(\overline{UP}) = 1 \quad (2)$$

Using Bayesian formula to expand the probability of negative samples in the task of "click-button-to-like", we have formula:

$$P(\overline{CBL}) = 1 - P(CBL) = P(\overline{CBL}|UP) * P(UP) + P(\overline{CBL}|\overline{UP}) * P(\overline{UP}) \quad (3)$$

Based on the above formulas, we can express our hidden objective. With the similar Bayesian expansion, we could rewrite the formulas for other tasks, in which single user's state P(UP) is shared among.

In prediction, we use formula (4) to predict the probability user prefer and click-button-to-like a feed.

$$P(CBL, UP) = P(CBL|UP) * P(UP) \quad (4)$$

Combined with the actual business, we can also find a hidden prior information: when the user does not prefer a feed, the probability of interaction is much smaller than the probability of interaction if the user prefers it. Hence, we can construct the following mathematical relationship:

$$P(CBL|UP) = \frac{N_{CBL}^{UP}}{N^{UP}} \gg P(CBL|\overline{UP}) = \frac{N_{CBL}^{\overline{UP}}}{N^{\overline{UP}}} \quad (5)$$

where $N_{CBL}^{UP}$ denotes the Number of users who prefer a feed and perform click-button-to-like, $N^{UP}$ denotes the number of users who prefer a feed, $N_{CBL}^{\overline{UP}}$ denotes the number of users who do not prefer an feed but perform click-bottom-to-like, $N^{\overline{UP}}$ denotes the number of users who do not prefer a feed.

It's easily proven that $\frac{N_{CBL}^{UP}}{N^{UP}} > \frac{N_{CBL}^{UP}+N_{CBL}^{\overline{UP}}}{N^{UP}+N^{\overline{UP}}} > \frac{N_{CBL}^{\overline{UP}}}{N^{\overline{UP}}}$ (6), so we can get a constraint that $0 < P(CBL|\overline{UP}) < \alpha$ (7), and the same goes for any other tasks. This result shows that there is an upper limit on the interaction probability when the user does not prefer it. According to our experience, the training effect of the model is better when the upper limit is set to 0.1~0.5 times the corresponding task. In the training phase, The Bayesian formulas can be understood as follows:

● When a user clicks "like" button, the model tends to increase both $P(CBL|UP)$ and $P(UP)$ because $P(CBL|\overline{UP})$ is very small.

● When the user doesn't click "like" button and there is no other interaction, the model tends to increase $P(\overline{UP})$ because $P(\overline{CBL}|UP)$ is very large.

● When the user doesn't click "like" button, but there are other positive interactions, the target of other positive interactions tends to increase $P(UP)$, which means to decrease the value of $P(\overline{UP})$. At the same time, the target user "prefer" will tend to increase the value of $P(\overline{CBL}|UP)$.

The third situation is what we should concern. The values of $P(\overline{CBL}|UP)$ and $P(\overline{UP})$ will be played in this case, and the overall result of the game will be determined by the sample distribution. Benefiting from this game process, we can guide the model learn the latent emotion state of the user $P(UP)$, which is used to describe the probability that the user favorites this feed.

### 3.2 Model structure

By combining the theory with reference to MMOE model, we propose Deep Latent Emotion Network (DLEN) model. The structure of model is shown in the figure. Unlike MMOE model, we have a separate output of hidden state network as the predicted value of $P(UP)$, which is shared among different objectives.

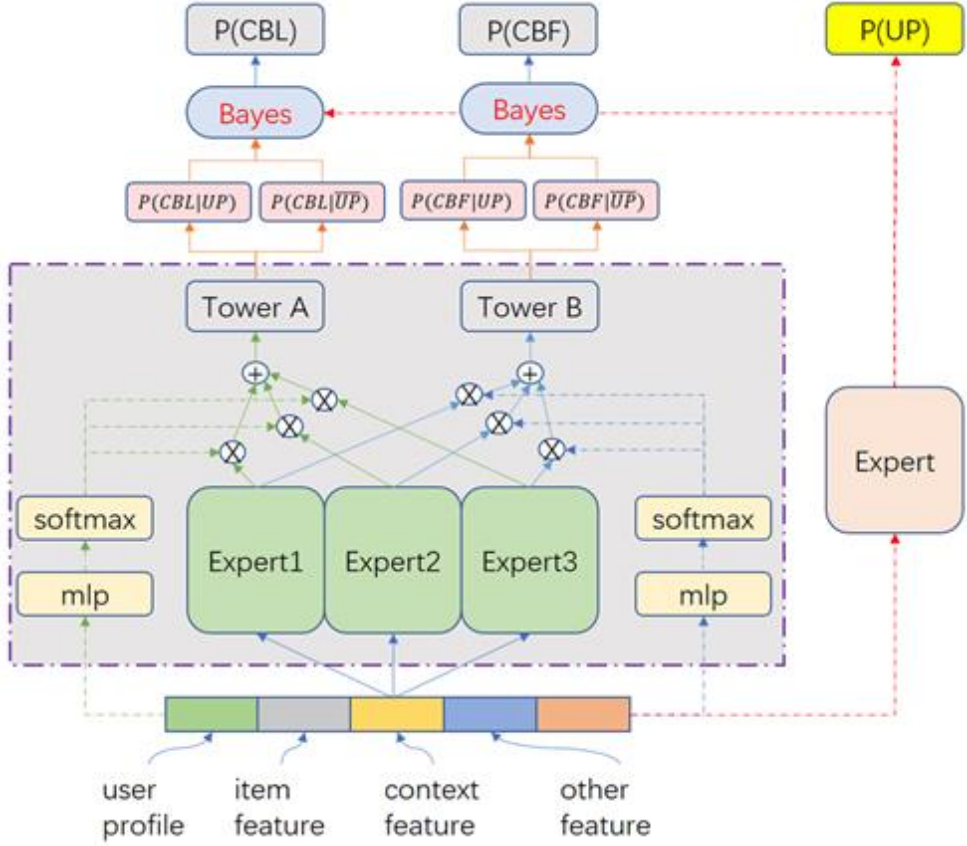

**Figure 2: Deep Latent Emotion Network**

In figure 2, $P(CBF)$ denotes the probability of click-button-to-follow, $P(CBF|UP)$ denotes the probability of user-prefer and click-button-to-follow, $P(CBF|\overline{UP})$ denotes the probability of user-not-prefer and click-button-to-follow.

Each tower outputs two values, Take CBL target as an example,

$$X^{CBL} = MMOE^{tower\_CBL}(X) \quad (8)$$
$$P(CBL|UP) = W^{CBL|UP}(X^{CBL}) \quad (9)$$
$$P(CBL|\overline{UP}) = \alpha * W^{CBL|\overline{UP}}(X^{CBL}) \quad (10)$$

where $X$ is the input feature of sample, $X^{CBL}$ is tower-CBL's last shared network's output of $P(CBL|UP)$ and $P(CBL|\overline{UP})$, $X^{CBL}$ has the sample calculation method in MMOE, $\alpha$ is a hyperparameter mentioned in 3.1, the final target CBL is calculated by formula 1 and other targets are calculated similarly. Besides, we use cross entropy as the loss function.

### 3.3 Evaluation metrics

In order to evaluate the training effectiveness of the model, we construct a relevant evaluation metrics. For ordinary integration behaviors, we use label and prediction score corresponding to the interaction to calculate AUC. For the hidden target, we can't evaluate it directly since the real information about whether the user favor a feed is not available. However, we could tell from figure 2 that user-interaction is a subset of user-favor, and when the predicted value of user-favor has a distinction on the objective of user liking an feed, it also has a certain degree of distinction on the objective of user-interaction. This is a necessary but not sufficient condition.

Based on this point, we use output of user-favor and label of user-interaction to calculate AUC as the metrics whether the evaluation model is converged. It should be noticed that this metrics is only used to qualitatively evaluate whether the output of user-favor is converged to our expected region, and not to quantitatively evaluate how well the output of user-favor converges.

## 4 EXPERIMENT

In this section, we conduct offline experiments and online ab-test comparison experiments on the large-scale Tencent feed dataset as well as the comparison experiments on the offline dataset of Ali-CPP. We compare and analyze the MMOE model and CGC model, which are used to verify the effectiveness of the user DLEN model.

### 4.1 Evaluation on Feed Recommendation System in Tencent

We used online data from Tencent Small World as our training dataset.

In this subsection, we conduct offline and online experiments for multiple tasks in Small-World at Tencent QQ to evaluate the effectiveness of DLEN model.

*4.1.1 Dataset.*

We collect online data from Tencent-Small-World as our training dataset. The Small-World is a feed recommendation business in Tencent QQ. The dataset we use in training baseline MTL model and DLEN model contains online real data of nearly 1 billion samples from Small-World. The dataset includes five tasks: click, like, push, follow, and comment.

In this experiment, we compared the DLEN model with the SOTA MTL model including MMOE and PLE.

The five tasks are all binary classification task, using cross entropy as the loss function for training, CTR as the prediction result, and AUC as the evaluation index of CTR.

For MMOE, we use 5 expert networks, and each expert network consists of a three-layer MLP network. Their hidden sizes are [256, 128, 64] respectively, using RELU as the activation function.

For the PLE model, in addition to 5 shared expert networks, each Task has 2 unique Expert networks. Each Expert network is the same as the MMOE model, which is also composed of a three-layer MLP network. The hidden size and activation function are also [256, 128, 64] and RELU.

For the PSE model, there are 5 Expert networks that are the same as the MMOE model. There is also a hidden state network, which has a three-layer MLP structure, their hidden size is [256, 128, 64], and the activation function is RELU.

*4.1.2 Models setup.*

In this experiment, we compared the DLEN model with the SOTA MTL model including MMOE and PLE model. The five tasks we mentioned before are all binary classification task trained with cross entropy as the loss function. The CTR is adopted as the prediction result with the evaluation index AUC. For the MMOE model, we use 5 expert networks, and each expert network consists of a three-layer MLP network. Their hidden sizes are [256, 128, 64] respectively and activation function is RELU. In addition to 5 shared expert networks, the PLE model has 2 unique Expert networks for each Task. Each Expert network is the same as the MMOE model, which is also composed of a three-layer MLP network, whose hidden size and activation function are also [256, 128, 64] and RELU. For the DLEN model, there are 5 Expert networks that are identical to the MMOE model. In addition, it has a hidden state network, which includes a three-layer MLP structure of hidden size [256, 128, 64], and the activation function RELU.

Refer to the paper[8], we use MTL gain to measure the benefit of each indicator relative to the base model MMOE. The offline experiment results are shown in the table 1:

Table 1: Performance on offline dataset

|  | AUC Click | AUC Like | AUC Follow | AUC Push | AUC Comment |
|---|---|---|---|---|---|
| MMOE | 0.7516 | 0.8937 | 0.9512 | 0.8679 | 0.8919 |
| PLE | 0.7525(+0.0009) | 0.8952(+0.0015) | 0.9514(+0.0002) | 0.8721(+0.0042) | 0.8956(+0.0023) |
| DLEN | **0.7532(+0.0016)** | **0.8971(+0.0034)** | **0.9519(+0.0007)** | **0.8747(+0.0068)** | **0.8948(+0.0029)** |

The offline experiment results show that the DLEN model is better than the SOTA MTL model: MMOE model and PLE model in different training tasks.

We deployed these models to the actual production environment, allocated the same amount of traffic for comparison experiments. Through one-month long-term tracking, we found that our DLEN model is also better than the Base model in terms of key indicators, user stay time and user page views.

Table 2: Improvement over base model on online A/B Test

|  | Click Rate | Like Rate | Follow Rate | Comment Rate | Stay time | PV |
|---|---|---|---|---|---|---|
| DLEN | +1.2% | +3.66% | +0.7% | 1.8% | +2.63% | +3.02% |

*4.1.3 Experiment results*

The offline experiment results demonstrate that both MTL model perform fairly in all tasks, where PLE model presents a better performance than MMOE in all metrics. In this scenario, the DLEN model has significant advantage in MTL gain to both MMOE and PLE model in all tasks, which reveals the benefit of DLEN model in multi-task learning.

Finally, we deployed the three models in a more challenging real-world production environment, with same amount of traffic allocated to comparison experiments. Through one-month long-term tracking, we found that the DLEN model presents a significant advantage in key indicators, user stay time and Page Views (PV) compared with baseline model MMOE.

**4.2 Evaluation on Ali-CCP dataset**

In this subsection, we conduct offline experiment on public dataset of Ali-CCP to verify the performance of proposed DLEN model.

*4.2.1 Dataset.*

Ali-CCP is a public dataset with 84 million samples provided by Alimama gathered from real-world traffic logs of the recommendation system in Taobao. This dataset contains only two targets: click and conversion. Therefore, our model is aiming at predicting the user's click-through rate CTR and user's conversion rate CVR, respectively.

Compared with experiment 4.1, we adjust the number of experts of the MMOE model to 2, and the hidden size to [256, 128, 64]. For the PLE model, the number of share experts is 2, the number of experts for each task is 1, and the hidden size of each expert is adjusted to [256, 128, 64].The DLEN model has also been modified accordingly, the number of experts is adjusted to 2, and the hidden size is adjusted to [256, 128, 64].

Other super participation in experiment 4.1 is consistent.

*4.2.2 Model setup.*

In this experiment, we make some adjustment in model's structure. Compared with experiment in 4.1, we adjust the number of experts in MMOE model to 2 and hidden size to [256,128,64]. For the PLE model, the number of share experts is changed to 2, and the number of experts for each task is 1, of which the hidden size is adjusted to [256,128,64]. The DLEN model is also been modified accordingly, where the number of experts is adjusted to 2, of which the hidden size is altered to [256,128,64]. Besides, the hyperparameters are consistent with 4.1 experimental hyperparameters.

Results on the Ali-CCP dataset are shown in the table 3.

Table 3: Performance on Ali-CPP dataset

|  | AUC CTR | AUC CVR |
|---|---|---|
| MMOE | 0.6231 | 0.6160 |
| MMOE-DLEN | **0.6245(+0.0014)** | **0.6282(+0.0122)** |
| PLE | 0.6241 | 0.6183 |
| PLE-DLEN | **0.6243(+0.0002)** | **0.6243(+0.0060)** |

The result shows that our DLEN model is better than SOTA MTL model：MMOE and PLE on CTR and CVR tasks.

*4.2.3 Experiment result*

The experiment results on Ali-CCP dataset are shown in the table. Compared with the SOTA MTL model MMOE and PLE, our DLEN model have better performance on both CTR and CVR tasks, which verifies DLEN model's effectiveness can be well generalized to other multi-tasks learning.

**4.3 Experiment analysis**

We conduct offline and online experiments on the Tencent information flow business dataset, and the results demonstrate that the proposed DLEN model outperforms both the MMOE model and PLE model in terms of all metrics in offline test. It also outperforms the base model in long-term metrics in the online experiment. The result of experiment generalized to the open dataset Ali-CCP also verifies that the DLEN model can obtain better effectiveness than MMOE and PLE model on both the CTR and CVR tasks.

## 5 CONCLUSION

This paper proposes a novel MTL model DLEN. We recognize that the bias generated by the complex composition of negative samples and the natural differences between tasks in the feed recommendation are the root causes of conflicts in multi-task models learning. Therefore, in DLEN we introduce Bayesian full probability formula into a neural-based MTL model, which decomposes the label of original multi-tasks, and introduces an auxiliary task to directly model the user's emotional tendency. In the training phase, this model reduces the difference between tasks and negative sample bias, thereby effectively alleviating the conflicts in model learning. In the prediction phase, the new target can play a positive role to improve the user experience. Our comparison experiments show that our DLEN models are all better than state-of-the-art MTL models. In the future, it will be the focus of our work to explore the way of fusion among different targets.